\documentclass[runningheads]{llncs}
\usepackage{amsmath}
\usepackage{amssymb}
\usepackage{cite}
\usepackage{graphicx}
\usepackage{grffile}
\usepackage[pdftex]{hyperref}
\usepackage{subfig}
\usepackage{tabularx}
\usepackage{multirow}

\graphicspath{{figures/}}

\begin{document}

% Title page
\mainmatter

\title{Self-Supervised Learning for Cardiac MR Image Segmentation by Anatomical Position Prediction}
\titlerunning{Self-Supervised Learning for Cardiac MR Image Segmentation}

\author{Wenjia Bai\inst{1,2} \and Chen Chen \inst{3} \and Giacomo Tarroni \inst{3} \and \\
Jinming Duan \inst{3} \and Florian Guitton \inst{1} \and Steffen E. Petersen\inst{4} \and \\
Yike Guo \inst{1} \and Paul M. Matthews \inst{2,5} \and Daniel Rueckert \inst{3}
}
% index{Bai, Wenjia}
% index{Chen, Chen}
% index{Tarroni, Giacomo}
% index{Duan, Jinming}
% index{Guitton, Florian}
% index{Petersen, Steffen E.}
% index{Guo, Yike}
% index{Matthews, Paul M.}
% index{Rueckert, Daniel}
\authorrunning{W. Bai et al}
\institute{
  Data Science Institute, Imperial College London, London, UK
  \and
  Department of Medicine, Imperial College London, London, UK
  \and
  BioMedIA, Department of Computing, Imperial College London, London, UK
  \and
  NIHR Barts BRC, Queen Mary University of London, London, UK
  \and
  UK Dementia Research Institute, Imperial College London, London, UK
}

\maketitle

% Abstract
\begin{abstract}
In the recent years, convolutional neural networks have transformed the field of medical image analysis due to their capacity to learn discriminative image features for a variety of classification and regression tasks. However, successfully learning these features requires a large amount of manually annotated data, which is expensive to acquire and limited by the available resources of expert image analysts. Therefore, unsupervised, weakly-supervised and self-supervised feature learning techniques receive a lot of attention, which aim to utilise the vast amount of available data, while at the same time avoid or substantially reduce the effort of manual annotation. In this paper, we propose a novel way for training a cardiac MR image segmentation network, in which features are learnt in a self-supervised manner by predicting anatomical positions. The anatomical positions serve as a supervisory signal and do not require extra manual annotation. We demonstrate that this seemingly simple task provides a strong signal for feature learning and with self-supervised learning, we achieve a high segmentation accuracy that is better than or comparable to a U-net trained from scratch, especially at a small data setting. When only five annotated subjects are available, the proposed method improves the mean Dice metric from 0.811 to 0.852 for short-axis image segmentation, compared to the baseline U-net.
\end{abstract}

% Main body
\section{Introduction}
Cardiac MR image segmentation plays a central role in characterising the structure and function of the heart. Quantitative phenotypes derived from the segmentations provide important biomarkers for diagnosing and managing cardiovascular diseases. In the recent years, convolutional neural networks have greatly advanced the performance of cardiac MR image segmentation due to their capacity in learning discriminative image features for the segmentation task \cite{Bernard2018, Bai2018,Tao2019}. Most successful methods are fully supervised and rely on a large amount of annotated data to learn the features. However, annotated medical imaging data may not always be available. The annotations are expensive to acquire and often limited by the available resource of expert image analysts. To address this challenge, we propose a novel way for training a cardiac MR image segmentation network, which formulates a self-supervised task for feature learning and alleviates the cost of data annotation.

There are two major contributions of this work. First, the proposed method learns image features from anatomical positions automatically defined by cardiac chamber view planes, which is a novel pretext task for self-supervised learning and provides a strong supervisory signal. Importantly, the chamber view plane information is freely available from standard cardiac MR scans, which means the method has the potential to be extended to a clinical setting, where a lot of unannotated MR scans stored on the PACS in hospitals can be utilised for feature learning. Second, we demonstrate the learning performance on two tasks, namely short-axis image and long-axis image segmentations. For both tasks, self-supervised learning demonstrates a strong boost to segmentation accuracy, especially at a small data setting.

\subsubsection{Related Works:}
To address the challenge of limited data annotations, there is increased interest in developing methods that do not require a large amount of annotations for feature learning. Directions of research include transfer learning, domain adaptation, semi-supervised, weakly-supervised, unsupervised and self-supervised learning \cite{Doersch2017}. Here, we focus on self-supervised learning, which formulates a \textit{pretext} task based on unannotated data for feature learning.

For natural image and video analysis problems, a number of pretext tasks have been explored, including prediction of image rotation \cite{Gidaris2018}, relative position \cite{Doersch2015}, colorisation \cite{Zhang2016} and image impainting \cite{Pathak2016} etc. In medical imaging domain, self-supervised learning has also been explored but to a less extent. Jamaludin et al. proposed a pretext task for subject identification \cite{Jamaludin2017}. A Siamese network was trained to classify whether two spinal MR images came from the same subject or not. The pretrained features were used to initialise a disease grade classification network. Ross et al defined re-colourisation of surgical videos as a pretext task and used the pretrained features to initialise a surgical instrument segmentation network \cite{Ross2018}. Tajbakhsh et al. used rotation prediction as a pretext task and the self-learnt features were transferred to lung lobe segmentation and nodule detection tasks \cite{Tajbakhsh2019}. Different from previous works in the medical imaging domain, we propose a novel pretext task, which is to predict anatomical positions. In particular, we leverage the rich information encoded in the cardiac MR scan view planes and DICOM headers to define the anatomical positions for the task.

\section{Methods}
Here we describe the cardiac MR view planes, the pretext task for self-supervised learning and architectures for transferring a self-trained network to a new task.

\begin{figure}[htb!]
  \centering
  \subfloat[Cardiac MR view planes]{
    \includegraphics[width=4.5cm]{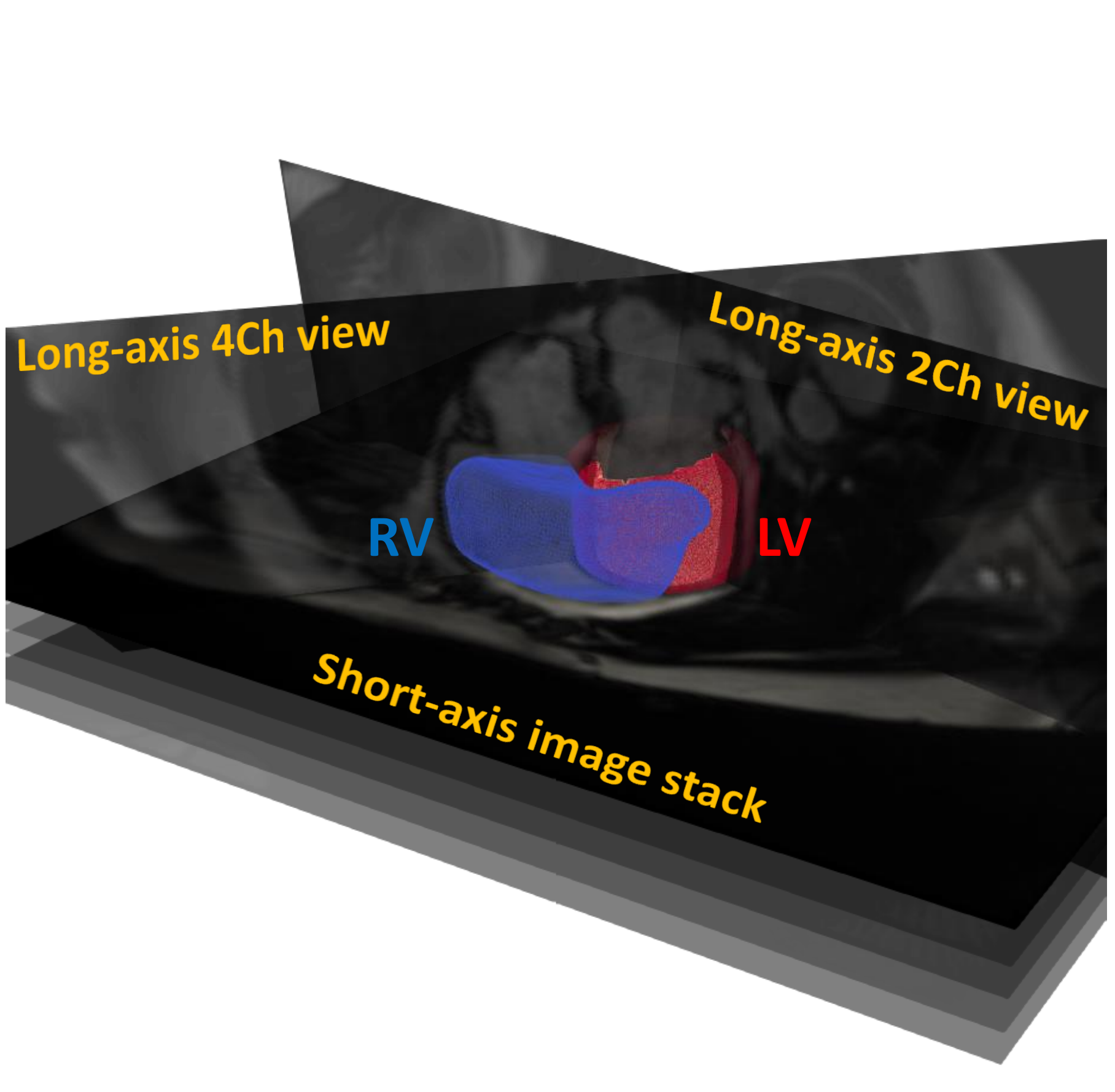}
  }
  \subfloat[Short-axis image]{
    \includegraphics[width=3.25cm]{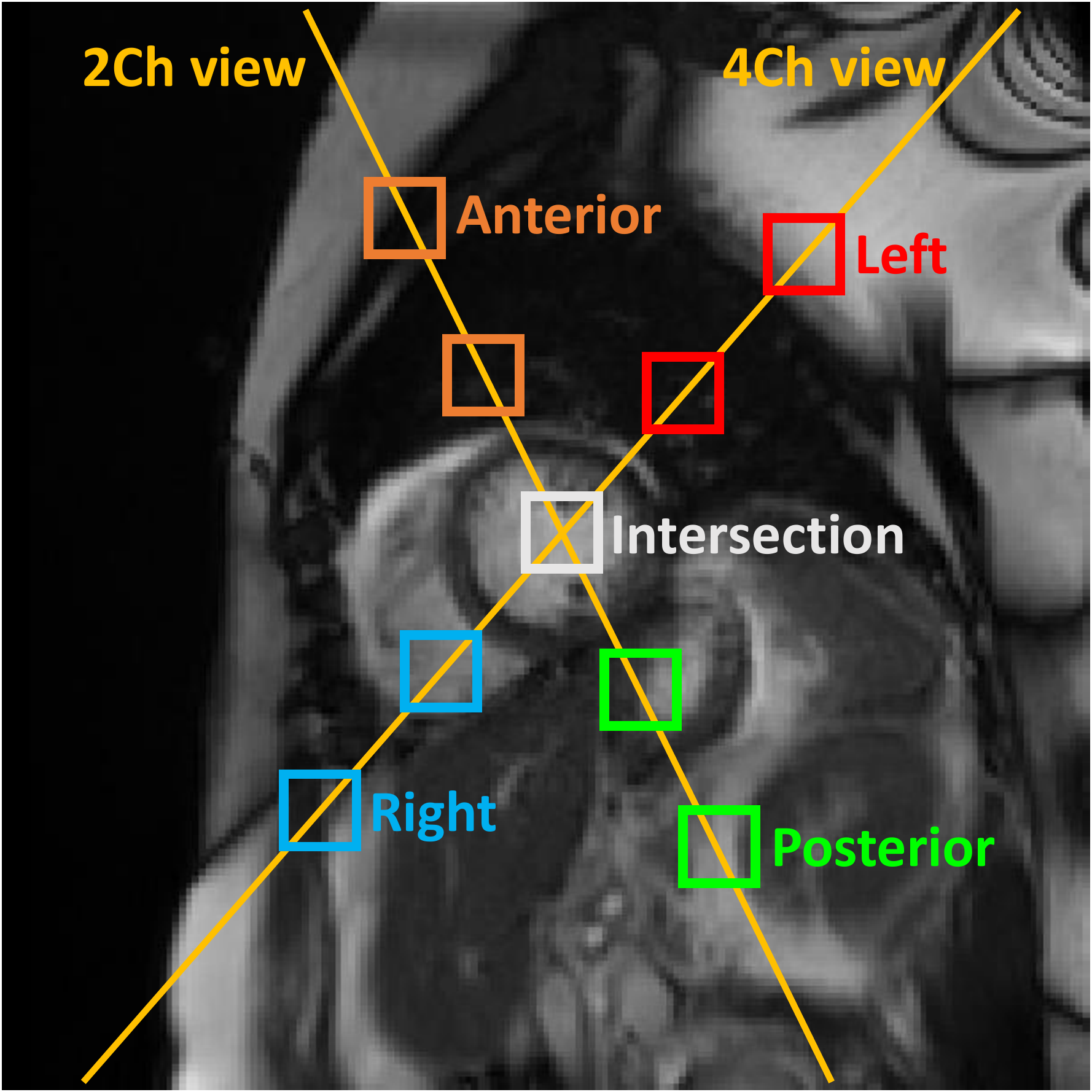}
  }
  \subfloat[Long-axis 4Ch image]{
    \includegraphics[width=3.25cm]{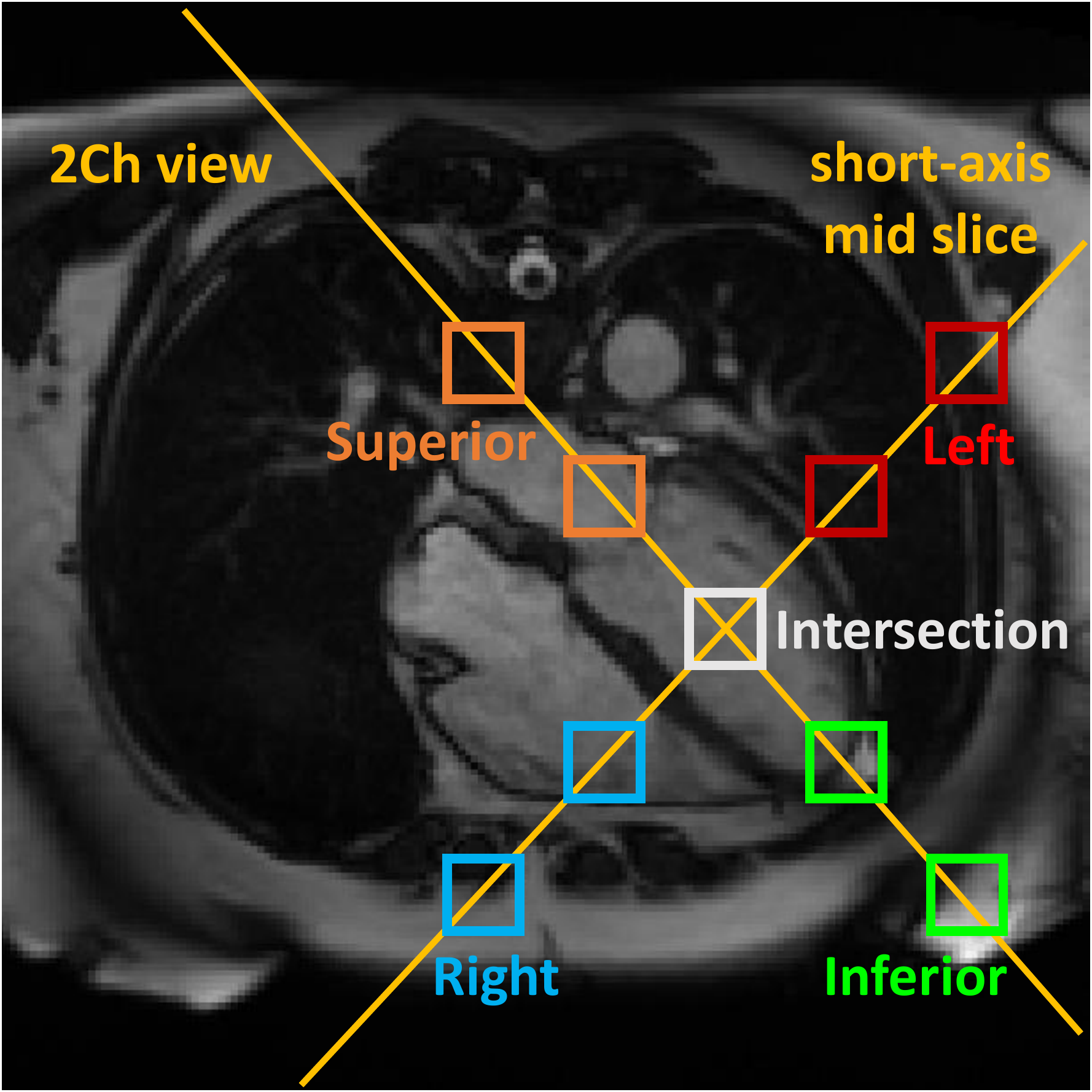}
  }
  \caption{Cardiac MR view planes and anatomical positions. (a) Short-axis and long-axis view planes with regard to the heart. (b) Short-axis image with overlaid 2Ch and 4Ch view planes (yellow lines) and anatomical positions defined by view planes (coloured boxes). (c) Long-axis 4Ch view image with overlaid 2Ch and mid short-axis view planes and anatomical positions. \label{fig:view}}
\end{figure}

\subsubsection{Cardiac MR view planes:} A standard cardiac MR scan consists of images acquired at different angulated planes with regard to the heart, including short-axis, long-axis 2 chamber (2Ch, vertical long-axis), 4 chamber (4Ch, horizontal long-axis) and 3 chamber (3Ch) views. They are used for evaluating different anatomical regions of the heart. For example, the short-axis view shows the cross-sections of the left ventricle (LV) and right ventricle (RV). The long-axis views show the septal and lateral walls of the ventricles, as well as the atrial chambers, including the left atrium (LA) and right atrium (RA). Figure~\ref{fig:view}(a) illustrates how the short-axis and long-axis 2Ch, 4Ch views are oriented with regard to the heart.

Most previous works on cardiac MR image segmentation \cite{Bernard2018, Bai2018,Tao2019} consider short-axis and long-axis views separately and disregard the relative orientation of different views. Typically, images from a specific view and the corresponding label maps are used to train a segmentation network from scratch. In this work, we propose that the relative orientation of short-axis and long-axis views and the anatomical positions, defined by the view planes, can be used to formulate a pretext task for training the network in a self-supervised manner and increasing data efficiency.

\subsubsection{Self-Supervised Learning (SSL):}
Figure~\ref{fig:view}(b) shows a short-axis image, with the overlaid 2Ch view and 4Ch view (yellow lines). As it shows, the 2Ch view bisects both the LV and RV, whereas the 4Ch view bisects the LV. They intersect at the LV. Along the chamber view lines, we define nine anatomical positions, represented by bounding boxes, including the intersection, two boxes on the left, two on the right, two at the anterior and two at the posterior. The orientations from left to right and from posterior to anterior are available from the DICOM headers. The pretext task is to predict the anatomical positions defined by these nine bounding boxes. The intuition here is that for the network to recognise these anatomical positions, it has to learn features for understanding not only where the left and right ventricles roughly are but also what their neighbouring regions look like. The learnt features can be transferred to a related but more demanding task, which is accurate segmentation of the ventricles.

\begin{figure}[htb!]
  \subfloat[SSL]{
    \includegraphics[width=2.57cm]{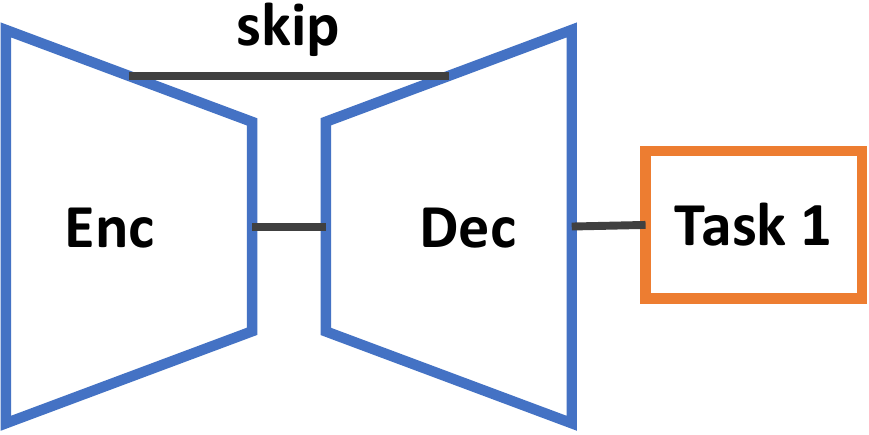}
  }
  \quad
  \subfloat[SSL+Decoder]{
    \includegraphics[width=2.57cm]{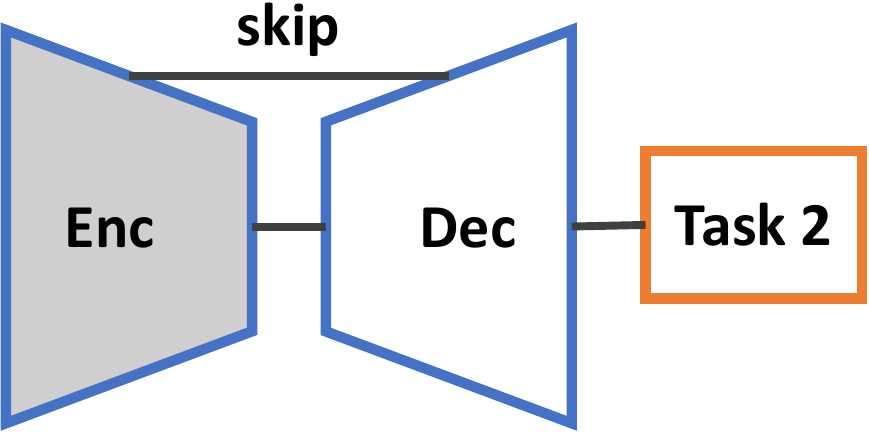}
  }
  \quad
  \subfloat[SSL+All]{
    \includegraphics[width=2.57cm]{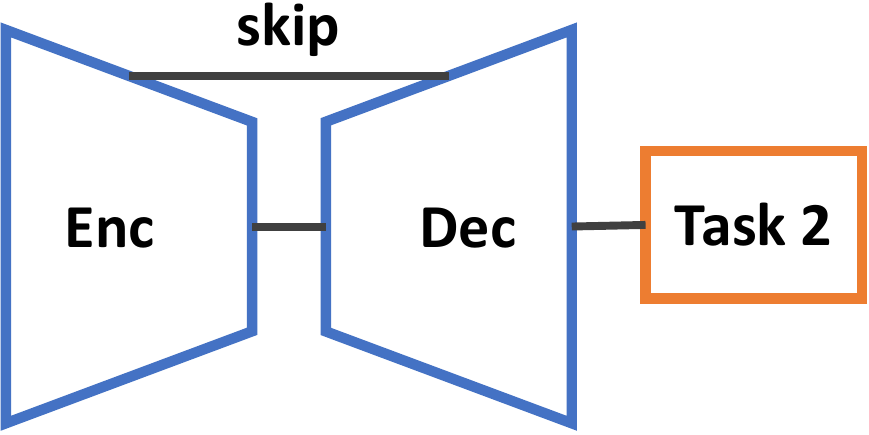}
  }
  \quad
  \subfloat[SSL+MultiTask]{
    \includegraphics[width=2.57cm]{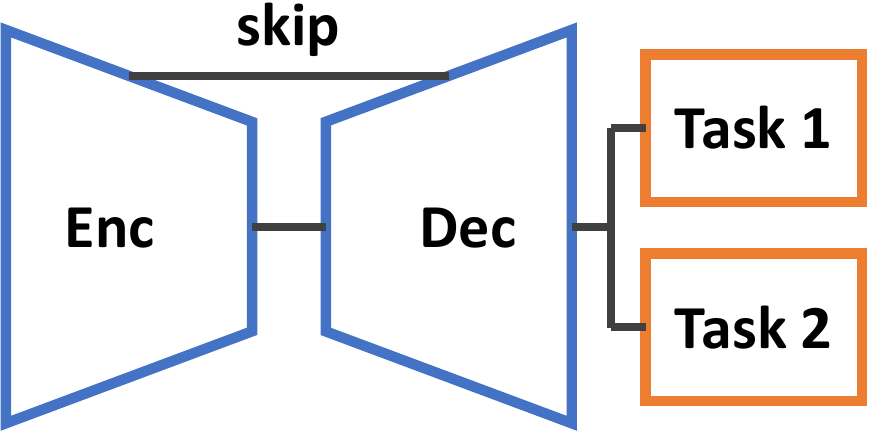}
  }
  \caption{Network architectures for self-supervised learning (SSL) and three different ways for transfer learning. The gray area in (b) denotes the freezed encoder. \label{fig:method}}
\end{figure}

Similarly, for a long-axis 4Ch view image, we can overlay the 2Ch view and mid short-axis view on it, shown by Figure~\ref{fig:view}(c). Along the 2Ch view and mid short-axis view lines, we define nine anatomical positions, including the intersection, two boxes on the left, two on the right, two at the superior and two at the inferior. The pretext task for long-axis image analysis is to predict these anatomical positions. To learn from the pretext task, we train a 10-way segmentation network, which segments the nine bounding boxes and the background. A standard U-net architecture \cite{Ronneberger2015} is used, which consists of the encoder part, decoder part, skip connections between them and the task head (the last convolutional layer), depicted by Figure~\ref{fig:method}(a). Cross-entropy is used as the loss function.

\subsubsection{Transfer Learning:}
After the network is self-trained on the pretext task (task 1), it is transferred to a new task (task 2), which is accurate segmentation of the anatomical structures, e.g. the LV cavity, myocardium and RV cavity. To achieve this, we can simply replace the head of task 1 by a new head for task 2. The task head refers to the last convolutional layer of the U-net, which is 1$\times$1 convolution with $K$-channel output, $K$ denoting the number of classes.

We investigate three different ways for transfer learning. The first way is to freeze the weights learnt at the encoder and only finetunes the decoder and task head, using the annotations for task 2. This method is named as ``SSL+Decoder'' and illustrated by Figure~\ref{fig:method}(b). The second way is to finetune all the weights \cite{Doersch2017}, including the encoder, decoder and task head. This method is named as ``SSL+All'' and illustrated by Figure~\ref{fig:method}(c). The third way is perform multi-task learning for finetuning. Two task heads are used, one for the pretext task and the other for the new task. This is to avoid forgetting about the pretext task while learning for the new task. This method is named as ``SSL+MultiTask'' and illustrated by Figure~\ref{fig:method}(d).  The loss function for multi-task learning is formulated as,
$L(\theta) = L_{task_1}(x,y_1|\theta) + \beta \cdot L_{task_2}(x,y_2|\theta)$,
where $\theta$ denotes the network parameters, $x$ denotes the image, $y_1$ denotes the label map of nine anatomical positions for task 1, $y_2$ denotes the label map of anatomical structures manually annotated by human experts for task 2, $\beta$ denotes the weight.

\section{Experiments and Results}
\subsubsection{Data:}
For self-supervised learning, short-axis and long-axis images of 3,825 subjects were used, acquired from the UK Biobank. The typical image dimension is 208$\times$180, with 1.82$\times$1.82 mm$^2$ in-plane resolution. The short-axis image stack consists of $\sim$10 slices. There is 1 slice for long-axis 4Ch view and 1 slice for 2Ch view. Bounding boxes were automatically placed at nine anatomical positions on the short-axis and long-axis 4Ch view images at end-diastole (ED) and end-systole (ES). Each box was empirically set to 11$\times$11 pixels and adjacent boxes were 30 pixels apart. For transfer learning, 200 subjects with manual annotations were used, which were randomly split into 100 subjects for training and 100 subjects for test. For short-axis images, LV, myocardium and RV were manually annotated by experienced image analysts at ED and ES frames. For long-axis 4Ch view images, LV, myocardium, RV, LA and RA were manually annotated.

\subsubsection{Implementation:}
The method was implemented using Tensorflow. For self-supervised learning, the Adam optimiser was used, with a learning rate of 0.001, a batch size of 20 image slices and 50,000 iterations. For transfer learning, the same setting was used. For both cases, data augmentation was performed online, including random rotation and scaling. For multi-task transfer learning, the weight $\beta$ was empirically set to 10 to emphasise the new task. As the input datasets were not consistent for the two tasks (task 1 with 3,825 subjects, task 2 with much fewer and different subjects), training was implemented as that at each iteration, task 1 was optimised for one sub-iteration, followed by optimising task 2 for $\beta$ sub-iterations. Since stochastic optimisation was performed, this was approximately equivalent to assigning a weight to task 2. Training multiple tasks alternately is a commonly adopted practice for inconsistent data input \cite{Doersch2017}. We made sure that task 2 was trained for 50,000 sub-iterations to enable a fair comparison. Finally, a U-net was also trained with exactly the same setting, but initialised with random weights. This is our baseline method, ``U-net-scratch''.

\begin{table}[htb!]
  \caption{Comparison of the Dice metrics for short-axis image segmentation. Column 1 lists the number of training subjects and manually annotated image slices. Columns 2 to 5 report the performance of the baseline method and different self-supervised learning methods. Values are mean (standard deviation). \label{tab:comp_sa}}
  \renewcommand{\arraystretch}{1.05}
  \centering
  \begin{tabular}{@{\quad}c@{\quad}c@{\quad}c@{\quad}c@{\quad}c}
    \hline
    \#subjects (\#slices) & U-net-scratch & SSL+Decoder & SSL+All & SSL+MultiTask \\
    \hline
    1 (18)  & 0.361 (0.047) & 0.515 (0.099) & 0.618 (0.068) & \textbf{0.704} (0.065) \\
    5 (102) & 0.811 (0.037) & 0.837 (0.048) & 0.844 (0.046) & \textbf{0.852} (0.046) \\
    10 (208) & 0.859 (0.037) & 0.860 (0.039) & 0.873 (0.036) & \textbf{0.875} (0.036) \\
    50 (980) & 0.876 (0.035) & 0.871 (0.038) & \textbf{0.884} (0.033) & 0.873 (0.037) \\
    100 (1,936) & 0.886 (0.030) & 0.867 (0.037) & 0.883 (0.035) & \textbf{0.887} (0.031) \\
    \hline
  \end{tabular}
\end{table}

\subsubsection{Short-axis image segmentation:}
We evaluated the performance on two transfer learning tasks, which are segmentations for short-axis and long-axis images. Table~\ref{tab:comp_sa} compares the Dice metrics (averaged across LV, myocardium and RV) for short-axis image segmentation between the U-net trained from scratch and self-supervised learning methods. As it shows, even if we freeze the encoder and only tune the decoder (SSL+Decoder), we can achieve a high accuracy comparable to training a U-net from scratch. This indicates that SSL is able to learn good features at the encoder which are transferrable for the segmentation task. The table also shows when we tune all the weights (SSL+All and SSL+MultiTask), the segmentation accuracy is generally better than U-net-scratch, especially when the number of training subjects is small. On average, SSL+MultiTask performs the best.

\begin{figure}[h!]
  \includegraphics[width=12cm]{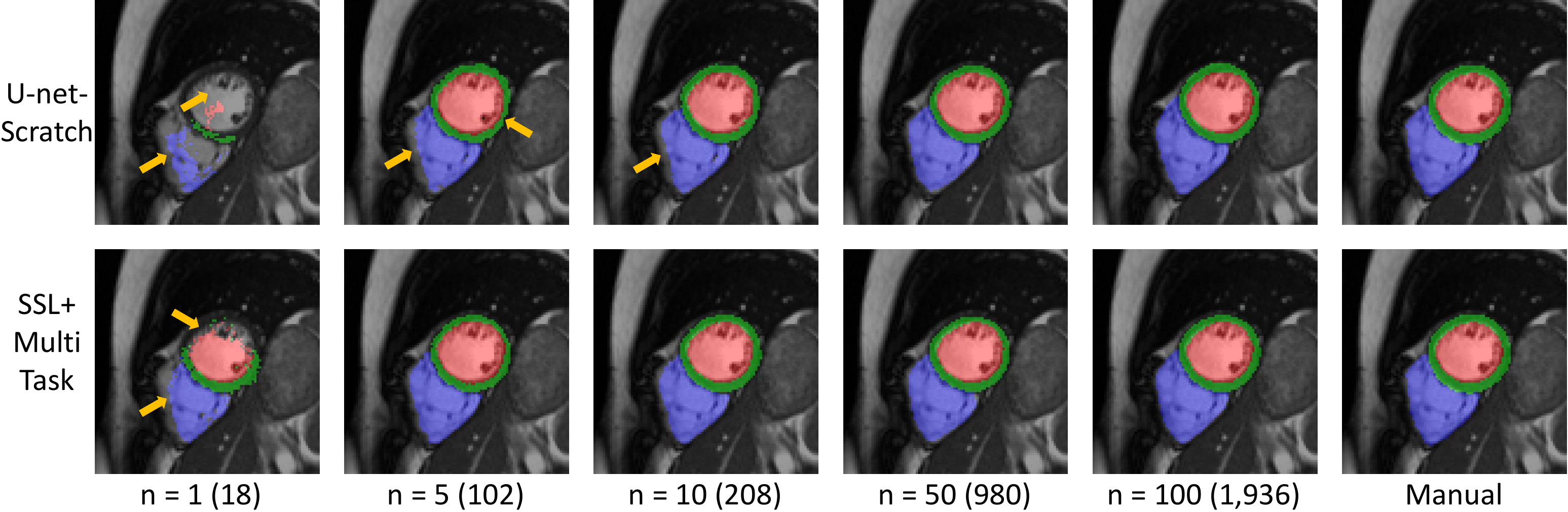}
  \caption{Short-axis image segmentations for U-net-scratch and SSL+MultiTask with an increasing number of training subjects (slices), as well as manual segmentations. The yellow arrows indicate segmentation errors. \label{fig:seg_comp_sa}}
\end{figure}

\begin{figure}[htb!]
  \centering
  \includegraphics[width=12cm]{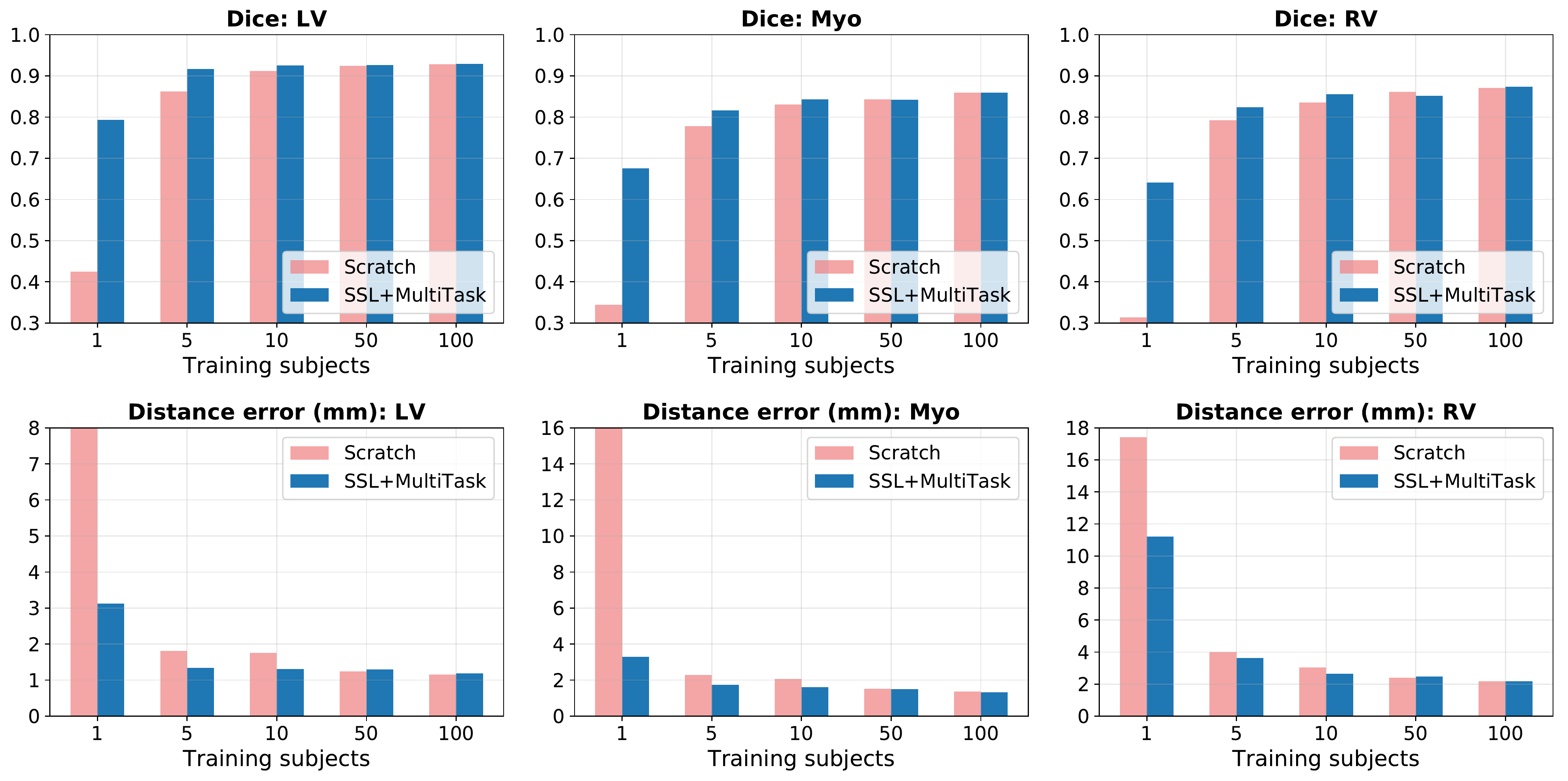}
  \caption{Comparison of the Dice metrics and mean contour distance errors on short-axis image segmentation for U-Net-scratch and SSL+MultiTask. \label{fig:plot_sa}}
\end{figure}

Figure~\ref{fig:seg_comp_sa} visualises exemplar segmentations for U-net-scratch and SSL+MultiTask with an increasing number of training subjects. It shows that when $n = 1$, due to the extremely small training set, U-net-scratch completely fails to segment the image. On the contrary, SSL+MultiTask is still able to segment some part of the LV and myocardium. When $n$ increases to 5 or 10, SSL+MultiTask outperforms U-net-scratch at details, for example, without RV under-segmentation errors. However, when $n$ increases to 50 or 100, the two methods perform similar to each other. Figure~\ref{fig:plot_sa} plots quantitative metrics including the Dice metric and mean contour distance error for each anatomical structure for the two methods. It shows a similar trend that at a small data setting ($n \le 10$), SSL+MultiTask outperforms U-net-scratch for all the structures. When there are more training data ($n \ge 50$), their performances become close to each other.

\begin{figure}[htb!]
  \includegraphics[width=12cm]{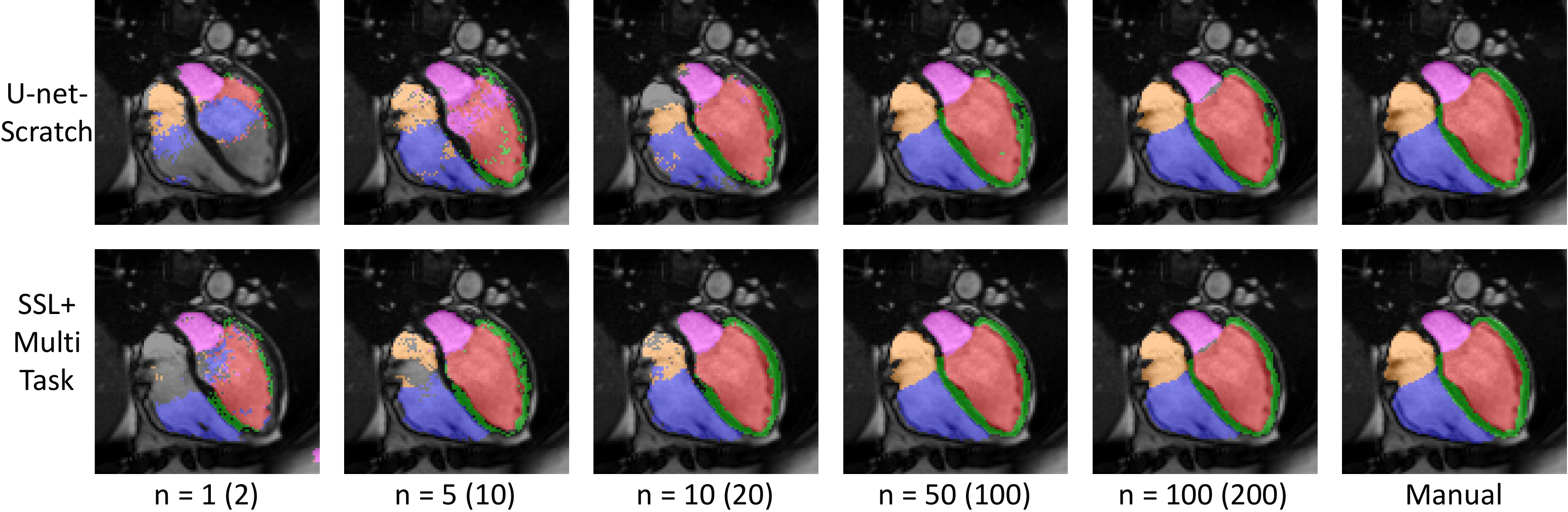}
  \caption{Long-axis image segmentations for U-net-scratch and SSL+MultiTask with an increasing number of training subjects (slices). \label{fig:seg_comp_la}}
\end{figure}

\begin{table}[htb!]
  \caption{Comparison of Dice overlap metrics for long-axis image segmentation. Values are mean (standard deviation). \label{tab:comp_la}}
  \renewcommand{\arraystretch}{1.05}
  \centering
  \begin{tabular}{@{\quad}c@{\quad}c@{\quad}c@{\quad}c@{\quad}c}
    \hline
    \#subjects (\#slices) & U-net-scratch & SSL+Decoder & SSL+All & SSL+MultiTask \\
    \hline
    1 (2) & 0.678 (0.132) & 0.699 (0.069) & \textbf{0.733} (0.081) & 0.600 (0.122) \\
    5 (10) & 0.848 (0.080) & 0.850 (0.068) & \textbf{0.875} (0.051) & 0.861 (0.049) \\
    10 (20) & 0.875 (0.044) & 0.888 (0.039) & \textbf{0.905} (0.039) & 0.889 (0.031) \\
    50 (100) & 0.922 (0.031) & 0.914 (0.035) & 0.925 (0.028) & \textbf{0.926} (0.029) \\
    100 (200) & 0.930 (0.032) & 0.924 (0.037) & 0.933 (0.031) & \textbf{0.934} (0.029) \\
	\hline
  \end{tabular}
\end{table}

\subsubsection{Long-axis image segmentation:}
We performed similar experiments for long-axis image segmentation. Table~\ref{tab:comp_la} reports the mean Dice overlap metrics. It shows that with SSL, for most of the cases, the segmentation accuracy is increased compared to U-net-scratch. Figure~\ref{fig:seg_comp_la} visualises exemplar segmentation results. It demonstrates that with limited training data, SSL+MultiTask generally produces better segmentations compared to U-net-scratch.

\section{Conclusions}
In this paper, we propose a novel method that leverages self-supervised learning for cardiac MR image segmentation. We formulate anatomical position prediction as the pretext task. Experiments on short-axis and long-axis image segmentation tasks demonstrate that with self-supervised learning, the proposed method outperforms a standard U-net trained from scratch at the small data setting and is of comparable performance at the large data setting. For future work, we will explore other anatomically meaningful pretext tasks to increase data efficiency in medical imaging applications.

\subsubsection*{Acknowledgements:}
This research has been conducted using the UK Biobank Resource under Application Numbers 2964 and 18545 and supported by the SmartHeart EPSRC Programme Grant (EP/P001009/1). We would like to thank NVIDIA Corporation for donating a Titan Xp for this research.

\bibliographystyle{unsrt}
\bibliography{refs}

\begin{thebibliography}{10}

\bibitem{Bernard2018}
O.~Bernard et~al.
\newblock {Deep learning techniques for automatic MRI cardiac multi-structures
  segmentation and diagnosis}.
\newblock {\em IEEE Trans Med Imaging}, 37(11):2514--2525, 2018.

\bibitem{Bai2018}
W.~Bai et~al.
\newblock {Automated cardiovascular magnetic resonance image analysis with
  fully convolutional networks}.
\newblock {\em J Cardiovasc Magn Reson}, 20(1):65, 2018.

\bibitem{Tao2019}
Q.~Tao et~al.
\newblock {Deep learning-based method for fully automatic quantification of
  left ventricle function from cine MR images}.
\newblock {\em Radiology}, 290(1):81--88, 2019.

\bibitem{Doersch2017}
C.~Doersch et~al.
\newblock {Multi-task self-supervised visual learning}.
\newblock In {\em ICCV}, 2017.

\bibitem{Gidaris2018}
S.~Gidaris et~al.
\newblock {Unsupervised representation learning by predicting image rotations}.
\newblock In {\em ICLR}, 2018.

\bibitem{Doersch2015}
C.~Doersch et~al.
\newblock {Unsupervised visual representation learning by context prediction}.
\newblock In {\em ICCV}, 2015.

\bibitem{Zhang2016}
R.~Zhang et~al.
\newblock {Colorful image colorization}.
\newblock In {\em ECCV}, 2016.

\bibitem{Pathak2016}
D.~Pathak et~al.
\newblock {Context encoders: Feature learning by inpainting}.
\newblock In {\em CVPR}, 2016.

\bibitem{Jamaludin2017}
A.~Jamaludin et~al.
\newblock {Self-supervised learning for spinal MRIs}.
\newblock In {\em MICCAI DLMIA Workshop}, 2017.

\bibitem{Ross2018}
T.~Ross et~al.
\newblock {Exploiting the potential of unlabeled endoscopic video data with
  self-supervised learning}.
\newblock {\em Int J Comput Assist Radiol Surg}, 13(6):925--933, 2018.

\bibitem{Tajbakhsh2019}
N.~Tajbakhsh et~al.
\newblock {Surrogate supervision for medical image analysis: Effective deep
  learning from limited quantities of labeled data}.
\newblock In {\em ISBI}, 2019.

\bibitem{Ronneberger2015}
O.~Ronneberger et~al.
\newblock {U-Net: convolutional networks for biomedical image segmentation}.
\newblock In {\em MICCAI}, 2015.

\end{thebibliography}

\end{document}